\documentclass{mva_style}

\usepackage{graphicx}
\usepackage{color}
\usepackage{comment}

\finalcopy 

\begin{document}
\title{Could you guess an interesting movie from the posters?:\\An evaluation of vision-based features on movie poster database}

\author{
  Yuta Matsuzaki$\dagger$ \and Kazushige Okayasu$\dagger$ \and Takaaki Imanari$\dagger$ \and Naomichi Kobayashi$\dagger$ \and Yoshihiro Kanehara$\dagger$ \and Ryousuke Takasawa$\dagger$ \and Akio Nakamura$\dagger$ \and Hirokatsu Kataoka$\ddagger$\\\\
  $\dagger$Department of Robotics and Mechatronics Tokyo Denki University\\
  5 Asahi-cho, Adachi-ku, Tokyo 120-8551, JAPAN\\
  {\tt \{matsuzaki.y, okayasu.k\}@is.fr.dendai.ac.jp}\\
  {\tt nkmr-a@cck.dendai.ac.jp}\\\\
  $\ddagger$National Institute of Advanced Industrial Science and Technology (AIST)\\
  1-1-1 Umezono, Tsukuba Ibaraki 305-8568, JAPAN\\
  {\tt hirokatsu.kataoka@aist.go.jp}\\
}



\maketitle

\section*{\centering Abstract}
\textit{ 
  In this paper, we aim to estimate the Winner of world-wide film festival from the exhibited movie poster. The task is an extremely challenging because the estimation must be done with only an exhibited movie poster, without any film ratings and box-office takings. In order to tackle this problem, we have created a new database which is consist of all movie posters included in the four biggest film festivals. The movie poster database (MPDB) contains historic movies over 80 years which are nominated a movie award at each year. We apply a couple of feature types, namely hand-craft, mid-level and deep feature to extract various information from a movie poster. Our experiments showed suggestive knowledge, for example, the Academy award estimation can be better rate with a color feature and a facial emotion feature generally performs good rate on the MPDB. The paper may suggest a possibility of modeling human taste for a movie recommendation.
}

\section{Introduction}

Imagine the situation when you see a movie at a theater. (see Figure 1).

Q. What if you decide the movie that you will watch?

A. Many movie posters are displayed in the theater, we will go to see the movie posters!

A good movie~\footnote{In the paper, we define a ``good movie" as a nominated movie for the world-wide film festival} have a visually interesting movie poster in advertisement. The benefit of movie poster is greater than as we expected. Elberse \textit{et al.} have placed that ``advertising has a positive and statistically significant impact on market-wide expectations prior to release" and ``the impact of advertising is lower for movies of lower quality" in the paper~\cite{1}. According to the statistics, a good movie poster seems to be instresting that has interesting contents in the movie. 

One one hand, the Academy award is one of the worldwide film festival which is explained as follows:

\vspace{-0.4\baselineskip}
\begin{itemize}
\item The Academy Awards, or ``Oscars", is an annual American awards ceremony hosted by the AMPAS to recognize excellence in cinematic achievements in the United States film industry as assessed by the Academy's voting membership.--(wikipedia)
\end{itemize}
\vspace{-0.3\baselineskip}

The Academy awards have a long history as much as the 89th in 2016. The successive movies remain in the memory and record. A good movie like nominated for the Academy awards focus a lot of efforts into advertisement. As for the poster, it effects a great contribution to the impact of the movie and its sales. Furthermore, the expert's eyes are refined in the long history, and it is excellent in a sensation to pick a good movie such a nominated one. Therefore, if we can learn a relationship between a winner and the poster, we enable to quantify the goodness of the movie from the poster.
Here, we introduce an award prediction in film festival and similar research in computer vision field.
\begin{figure}[t]
\begin{center}
   \includegraphics[width=0.78\linewidth]{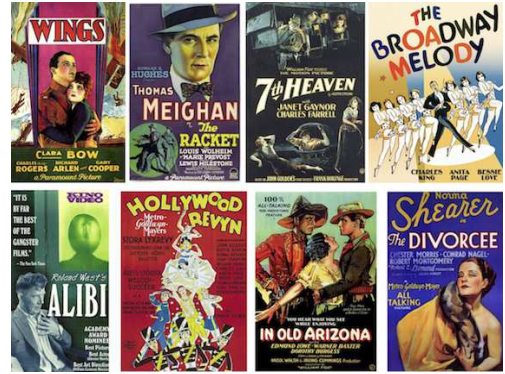}
   \caption{This is an example of an exhibited movie poster. \textit{Question:} Could you identify which movie from this movie poster is the film festival Winner? \protect \footnotemark We will address this issue.}
   \label{fig:flowchart_stdd}
\end{center}
\end{figure}
\footnotetext{\textit{Answer}: Wings, the top-left movie poster}
\textit{Award prediction} 
In the Academy award prediction, Krauss \textit{et al.} made a prediction based on a correlation value by utilizing a change of box-office takings, social situations and ratings on the IMDb~\footnote{http://www.imdb.com/}~\cite{2}. Although the accurate prediction is clear since the system enables to estimate an Academy award from various information, the approach is not helpful for choosing a good movie at a very moment. If we can model an expert's judgment with a movie poster, we instantly select a good movie at the movie theater.


\textit{Related works in CV} Joo \textit{et al.} succeeded to estimate election winning / losing from a facial image~\cite{3}. They extracted a low-level, mid-level feature from a face attribute and parts appearance of candidates and evaluated them. Based on the evaluation, it is possible to estimate that election winning / losing of candidates with 70 \% accuracy. Furthermore, even for the task of estimating camera-man from given pictures, Thomas \textit{et al.} have significantly succeeded an estimation with 75\% rate for 41 categories~\cite{4}．


Above these lines, we believe that the vision-based technology allows us to predict a winner in the world-wide film festival with movie posters. 


In this paper, we estimate a winner in the world-wide film festival from nominated movie posters. We have collected a database which contains a large amount of movie posters in the 4 biggest film festival (Academy, Berlin, Cannes, Venice). The movie poster database (MPDB) has historic movies over 80 years which are nominated a movie award at each year. We employ various types of features such as hand-craft (e.g. HOG, SIFT), mid-level (e.g. Face) and deep (e.g. AlexNet, VGGNet) features on the MPDB. In the experiments, we estimate a winner at each film festival with various features. 



The contributions of the paper are as follows.
\vspace{-0.4\baselineskip}
\begin{itemize}
\setlength{\itemsep}{-0.0mm}
\setlength{\parskip}{-0.0mm}
\item The estimation of winner in the 4 biggest film festival with movie posters. To tackle the problem, we have evaluated various types of features.
\item The collection of movie poster database (MPDB) which has 3,500+ nominates and 290+ winners over 80 years. 
\end{itemize}


\section{Movie poster database (MPDB)}
Posters of this dataset are collected from Internet Movie Database (IMDb). The MPDB is consist of the nominations and winner films in the 4 biggest film festivals (Academy Awards, Berlin International Film Festival, Cannes International Film Festival, Venice International Film Festival). The image database contains 3,844 nominates, 295 winners in the film festivals. The Berlin International Film Festival, the Cannes International Film Festival, and the Venice international Film Festival are the 3 notable film festivals in the world. The Academy award is an authoritative award with a history that is older than the three major film festivals, and its influence on the box-office takings of the awards and the economy. The structure and detail of the MPDB are shown in Table 1. The poster images are annotated the winners and nominates. The database has been collected on the IMDb which is a prestigious movie rating service. However, we did not collect any ratings and other captions.



\begin{table*}[t]
　 \caption{Constitution of dataset}
  \begin{center}
  \scalebox{0.9}{
  \begin{tabular}{|c|c|c|c|c|} 
  \hline
  Film Festival & year & \#winners & \#nominates & Ave. \#nominations \\
   &  &  &  & at each year \\ \hline \hline
  Academy & 1929-1932,1934-2016 & 88 & 440 & 16.6 \\
  Berlin & 1951-2016 & 63 & 905 & 6.50 \\
  Cannes & 1939,1946-1947,1949,1951-1968,1969-2016 & 91 & 1335 & 6.38 \\ 
  Venice & 1932,1934-1942,1946-1972,1979-2016 & 53 & 869 & 5.75 \\ \hline
  \end{tabular}
  }
  \end{center}
\end{table*}

\begin{figure}[t]
   \begin{center}
   \includegraphics[width=0.7\linewidth]{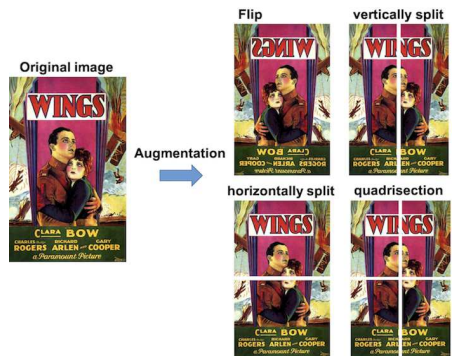}
   \caption{Data augmentation}
   \label{fig:augment}
   \end{center}
\end{figure}

In the settings of training and testing, we assign a leave-one-year-out-cross-validation at each film festival. When a certain year is tested for winner prediction, others for the training. In the winner prediction, we calculate a posterior probability of a movie poster to be identified, then let the highest probability is selected as a winner. In addition to the original database, the training data is also subjected to data augmentation with 4 divisions (see Figure~\ref{fig:augment}). In the Cannes and Venice, since there are two or more winner's works, in that case winner acquired works in order from the posterior probability top. Moreover, there is a year that there are 0 winner works, so Winner estimation is not done for that year.

\begin{figure}[t]
   \begin{center}
   \includegraphics[width=1.0\linewidth]{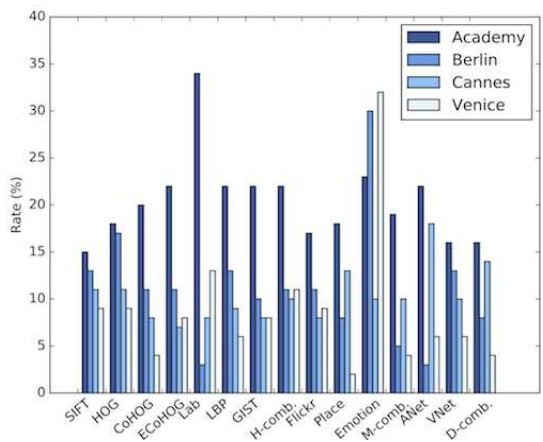}
   \caption{Comparison of various features in (i) handcraft (ii) mid-level (iii) deep features}
   \label{fig:rates}
   \end{center}
\end{figure}




\section{Features}
Our goal is to train and estimate an award winner from movie posters. We have evaluated various features for the task. We employed (i) hand-craft feature, (ii) mid-level feature (iii) deep feature in order to extract various information from the posters in the MPDB. We itemize the features as below.






\begin{description}
\setlength{\itemsep}{-0.4mm}
\setlength{\parskip}{-0.4mm}
\item[Handcraft feature]\mbox{}
  \begin{itemize}
  \setlength{\itemsep}{-0.0mm}
  \setlength{\parskip}{-0.0mm}
  \item SIFT + BoF\cite{5}\cite{6}
  \item HOG\cite{7}
  \item CoHOG\cite{8}
  \item ECoHOG\cite{9,9_2}
  \item L*a*b*\cite{10}
  \item LBP\cite{11}
  \item GIST\cite{12}
  \item Combined handcraft feature 
  \end{itemize}

\item[Mid-level feature]\mbox{}
  \begin{itemize}
  \setlength{\itemsep}{-0.0mm}
  \setlength{\parskip}{-0.0mm}
  \item FlickrStyle\cite{13}
  \item PlaceNet\cite{14}
  \item EmotionNet\cite{15}
  \item Combined mid-level feature
  \end{itemize}

\item[Deep feature]\mbox{}
  \begin{itemize}
  \setlength{\itemsep}{-0.0mm}
  \setlength{\parskip}{-0.0mm}
  \item AlexNet (DeCAF)\cite{16}
  \item VGGNet (DeCAF)\cite{17}
  \item Combined deep feature
  \end{itemize}


\end{description}

Lab expresses each expression of L,a,b as 30 bins. Emotion expresses by inputting the face image acquired with Haar-like feature to EmotionNet and voting on the histogram to which the expression bin is allocated (Each histograms normalized). Also, rough face position information is included in the histogram (upper-left, upper-right, lower-left, lower-right). Therefore, the dimension of the face feature acquired in this research is 32.
Classifier. The extracted feature quantity is identified by SVM. The parameters for identification were set to $C = 5.0 \times 10 ^ {4}$, $gamma = 1.0 \times 10 ^ {- 5}$, $kernel = "rbf"$.
Combine. We combine features extracted by each feature extraction method. As a method of combination, ``late fusion" is adopted.


\begin{figure*}[t]
   \begin{center}
   \includegraphics[width=0.82\linewidth]{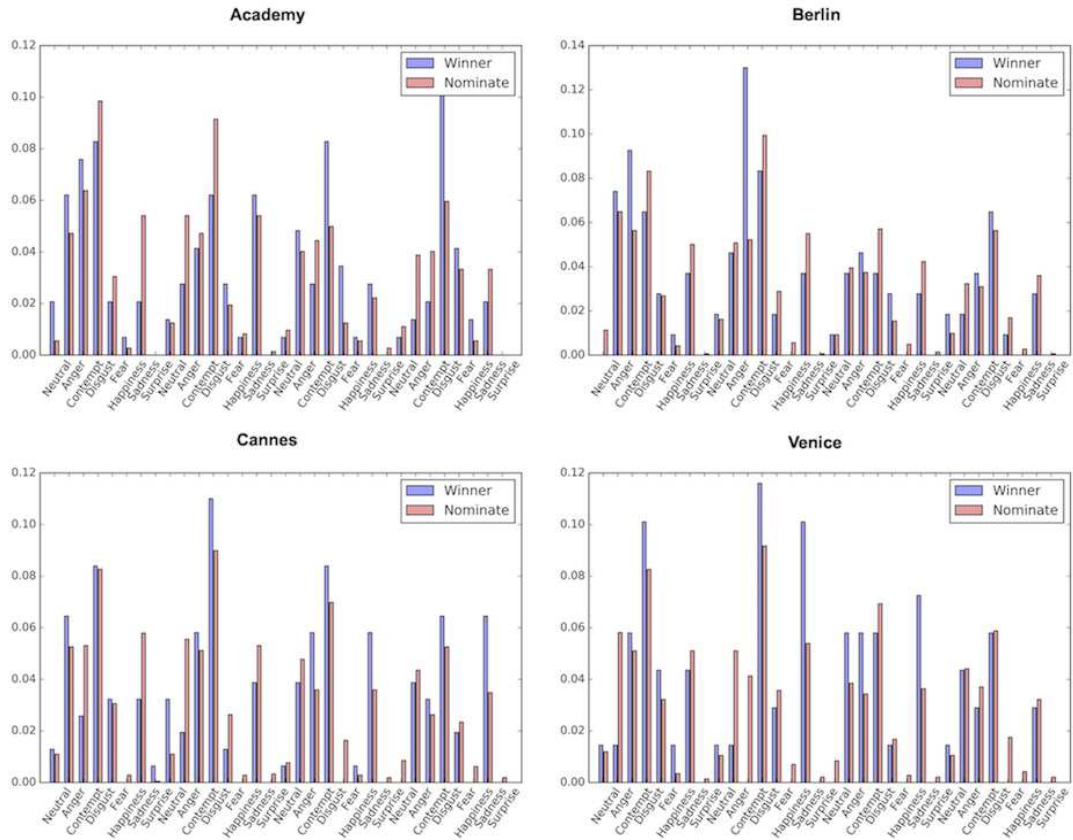}
   \caption{Facial expression histogram between winner and nominate for each film festival(Bins of expression are arranged in order of upper left, upper right, lower left, lower right).}
   \label{fig:facehist}
   \end{center}
\end{figure*}

\section{Exprimental Evaluation}
The results of the winner prediction are shown in Figure~\ref{fig:rates}. According to the Figure~\ref{fig:rates}, it was confirmed that the identification results by the L*a*b* feature was the highest in the Academy award and relatively high with the EmotionNet. In addition, the Berlin and Venice also showed that the recognition rate using EmotionNet expression is the highest identification rate. We will consider focused on the feature whose identification rate is 30 \% or more. In the Academy awards, there are many tendencies in which works selected as Winner tend to have certain colors. By analyzing L*a*b* histogram, we gained a knowledge that there are a couple of warm colors such as red, yellow, brown, etc., some of winners. We consider these colors contribute to improving the identification rate in the identification by the kernel SVM. In Figure~\ref{fig:facehist}, it can be seen that mode of facial expression is different in Berlin. In particularly Winner's works are noticeable `Contempt' in the upper left and upper right. In Venice (Figure~\ref{fig:facehist}), the remarkable difference is `Anger' in the upper-left, `Sadness' in the upper-right and lower-left.

Futher, it can be seen from figure~\ref{fig:facehist} that the expression histogram of Winner's works differs at each film festival. In Academy `Disgust' appears in lower-right. In Berlin, `Contempt' appears frequently in upper-right.In Cannes, `Disgust' appears in upper-right and `Sadness' appears in lower-right. In Venice, `Sadness' appears in upper-right more. From this result, we suggested that it is possible to predict the movie festival by facial expression, because the facial expression drawn on the poster is different for each movie festival.

In Cannes, accuracy was low in any feature, and there was no tendency of discrimination. We have considered that Cannes had chosen Winner based on box office income and the social situation at the time of movie release.

\section{Application (further result in Academy 2017)}
On February 26, 2017, the Academy Award 2017 was announced. We performed the winner prediction with our approach which assigns parameters from the experiments. The prediction results are shown in Table~\ref{tab:academy2017}. As shown in Table~\ref{tab:academy2017}, we have obtained interesting results at Academy Award 2017. Our system got a correct answer as the Academy Award 2017 ``moonlight".

\begin{table}[t]
　 \caption{Result of our prediction on Academy 2017 (sorted by predicted score)}
  \begin{center}
  \scalebox{0.9}{
  \begin{tabular}{|c|c|c|} 
  \hline
  Movie title & Predicted score & Rank \\ \hline \hline
  Moonlight [Winner] & 0.167 & 1 \\
  Lion & 0.163 & 2 \\
  Hell or High Water & 0.162 & 3 \\
  Arrival & 0.151 & 4 \\
  Hacksaw Ridge & 0.142 & 5 \\
  Fences & 0.138 & 6 \\
  Hidden Figures & 0.114 & 7 \\
  Manchester by the Sea & 0.112 & 8 \\
  La La Land & 0.093 & 9 \\
  \hline
  \end{tabular}
  }
  \label{tab:academy2017}
  \end{center}
\end{table}

\section{Conclusion}
In this work, we addressed a novel problem of estimating winner from a exhibited movie poster at film festival. In order to tackle this problem we have created a new dataset which is consist of winners and nominates in the 4 world-wide film festivals. The experiments on the self-collected MPDB (movie poster database) indicated a couple of suggestive realities such that the Academy award can be predictable with a color feature (L*a*b* corrects 34 \% even though the random is around 15 \%), also Berlin and Venice can be predictable with a emotional feature from face expression generally achieves well (around 25 \% in the Academy, Berlin and Venice film festivals) on the MPDB. In Cannes, accuracy was low in any feature, and there was no tendency of winner.

This paper suggested the possibility of modeling human intuitions when selecting a movie. In the future, we would like to create a more sophisticated DCNN framework to estimate a winner in a movie festival.


\begin{thebibliography}{99}


\bibitem{1}
  A. Elberse, et al.: 
  ``The effectiveness of pre-release advertising for motion picture: An empirical investigation using a simulated market,'' 
  \textit{Information Economics and Policy}, vol.19, pp.319--343, 2007.


\bibitem{2}
  J. Krauss, et al.:``Predicting Movie Success and Academy Awards Through Sentiment and Social Network Analysis,''
  \textit{In Proc .of 16th Europian Coference on Information Systems}, 2008.

\bibitem{3}
  J. Joo, et al.: 
  ``Automated Facial Trait Judgment and Election Outcome Prediction:
Social Dimensions of Face,'' 
  \textit{In Proc .of 2015 IEEE International Conference on Computer Vision}, 2015.

\bibitem{4}
  C. Thomas, et al.: 
  ``Seeing Behind the Camera: Identifying the Authorship of a Photograph,'' 
  \textit{Computer Vision and Pattern Recognition}, 2016.

\bibitem{5}
  D. G. Lowe, et al.: 
  ``Distinctive Image Features from Scale-Invariant Keypoints,'' 
  \textit{International Journal of Computer Vision}, vol.60, no.2, pp.91--110, 2004.

\bibitem{6}
  G. Csurka, et al.: 
  ``Visual Categorization with Bags of Keypoints,'' 
  \textit{Europian Coference on Computer Vision}, pp.1--22, 2004.

\bibitem{7}
   N. Dalal, et al.: 
  ``Histograms of Oriented Gradients for Human Detection,'' 
  \textit{Computer Vision and Pattern Recognition}, vol.1, no.1, pp.886--893, 2005.

\bibitem{8}
  T. Watanabe, et al.: 
  ``Co-occurrence Histograms of Oriented Gradients for Pedestrian Detection,'' 
  \textit{Proceedings of the 3rd Pacific Rim Symposium on Advances in Image and Video Technology}, pp.37--47, 2009.

\bibitem{9}
  K. Kataoka, et al.: 
  ``Symmetrical Judgment and Improvement of CoHOG Feature Descriptor 
for Pedestrian Detection,'' 
  \textit{International Conference on Machine Vision Applications}, 2011.

\bibitem{9_2}
  H. Kataoka, et al.:
  ``Extended Feature Descriptor and Vehicle Motion Model with Tracking-by-detection for Pedestrian Active Safety,"
  \textit{IEICE Transactions on Information and Systems}, Vol.E97-D, No.2, 2014.


\bibitem{10}
  F. Palermo, et al.: 
  ``Dating historical color images,'' 
  \textit{In Proceedings of the European Conference on Computer Vision}, pp.499--512, 2012.

\bibitem{11}
  T. Ojala, et al.: 
  ``A Comparative Study of Texture Measures with Classification Based on Feature Distributions,'' 
  \textit{Pattern Recognition}, vol.29, pp.51--59, 1996.

\bibitem{12}
  A. Oliva, et al.: 
  ``Modeling the Shape of the Scene: A Holistic Representation
of the Spatial Envelope,'' 
  \textit{International Journal in Computer Vision}, vol.13, no.42, pp.143--175, 2001.

\bibitem{13}
  S. Karayev, et al.: 
  ``Recognizing Image Style,'' 
  \textit{Proceedings of the British Machine Vision Conference}, 2013.

\bibitem{14}
  B. Zhou, et al.: 
  ``Learning Deep Features for Scene Recognition using Places Database,'' 
  \textit{Advances in Neural Information Processing Systems}, 2014.

\bibitem{15}
  B. Kennedy, A. Balint, 2016.\\
  Github: {\tt https://github.com/co60ca/EmotionNet}

\bibitem{16}
  A. Krizhevsky, et al.: 
  ``mageNet Classification with Deep Convolutional Neural Networks,'' 
  \textit{Advances in Neural Information Processing Systems}, 2012.

\bibitem{17}
  K. Simonyan, et al.: 
  ``Very Deep Convolutional Networks for Large-Scale Image Recognition,'' 
  \textit{International Conference on Learning Representations}, 2015.






\end{thebibliography}

\end{document}